\documentclass{article}

\usepackage{times}
\usepackage{xr,pifont}
\usepackage{mathtools}
\usepackage{graphicx} 
\usepackage{subfigure, caption} 
\usepackage{amsmath, amsthm, amssymb}
\usepackage{amsfonts}       
\usepackage{nicefrac}       
\usepackage{microtype}      
\usepackage{bm, stmaryrd}
\usepackage{enumitem} 
\usepackage{color}
\usepackage{natbib}

\usepackage{algorithm}
\usepackage{algorithmic}

\usepackage{hyperref}

\usepackage[accepted]{icml2018}

\usepackage{comment} 

\DeclarePairedDelimiter\floor{\lfloor}{\rfloor}

\newcommand{\bR}{\mathbb{R}} 

\newcommand{\cG}{\mathcal{G}}
\newcommand{\utilb}{\mathcal{G}}

\newcommand{\cN}{\mathcal{N}}

\newcommand{\cS}{\mathcal{S}}

\newcommand{\cV}{\mathcal{V}}
\newcommand{\cX}{\mathcal{X}}

\newcommand{\baa}{\bm{a}}
\newcommand{\bb}{\bm{b}}
\newcommand{\bc}{\bm{C}}
\newcommand{\bC}{\bm{C}}
\newcommand{\bE}{\mathbb{E}}
\newcommand{\bP}{\mathbb{P}}
\newcommand{\tp}{\text{TP}}
\newcommand{\fp}{\text{FP}}
\newcommand{\tn}{\text{TN}}
\newcommand{\fn}{\text{FN}}
\newcommand{\cU}{\mathcal{U}}
\newcommand{\util}{\mathcal{U}}

\newcommand{\heta}{\hat{\eta}}

\newcommand{\sign}[1]{\text{sign}\left({#1}\right)}
\newcommand{\innerprod}[2]{\langle {#1}, {#2}\rangle}

\newcommand{\bas}[1]{\begin{align*}#1\end{align*}}
\newcommand{\ba}[1]{\begin{align}#1\end{align}}

\definecolor{alizarin}{rgb}{0.82, 0.1, 0.26}
 
\newcommand{\vertiii}[1]{{\left\vert\kern-0.25ex\left\vert\kern-0.25ex\left\vert #1 
    \right\vert\kern-0.25ex\right\vert\kern-0.25ex\right\vert}}

\newcommand{\rom}[1]{%
  \textup{\uppercase\expandafter{\romannumeral#1}}%
}

\newtheorem{definition}{Definition}[section]
\newtheorem{theorem}{Theorem}[section]
\newtheorem{corollary}{Corollary}[section]
\newtheorem{assumption}{Assumption}
\newtheorem{lemma}{Lemma}[section]

\newtheorem{proposition}{Proposition}[section]

\icmltitlerunning{Binary Classification with Karmic, Threshold-Quasi-Concave Metrics}

\begin{document} 

\twocolumn[
\icmltitle{Binary Classification with Karmic, Threshold-Quasi-Concave Metrics}

\icmlsetsymbol{equal}{*}

\begin{icmlauthorlist}
\icmlauthor{Bowei Yan}{ut}
\icmlauthor{Oluwasanmi Koyejo}{uiuc}
\icmlauthor{Kai Zhong}{ut}
\icmlauthor{Pradeep Ravikumar}{cmu}
\end{icmlauthorlist}

\icmlaffiliation{ut}{University of Texas at Austin, Austin, Texas, USA}
\icmlaffiliation{uiuc}{University of Illinois at Urbana-Champaign, Champaign, Illinois, USA}
\icmlaffiliation{cmu}{Carnegie Mellon University, Pittsburgh, Pennsylvania, USA}

\icmlcorrespondingauthor{Bowei Yan}{boweiy@utexas.edu}

\icmlkeywords{classification, non-decomposable performance metrics, karmic metrics, Threshold-Quasi-Concave}

\vskip 0.3in
]
\printAffiliationsAndNotice{}

\begin{abstract}
Complex performance measures, beyond the popular measure of accuracy, are increasingly being used in the context of binary classification. These complex performance measures are typically not even decomposable, that is, the loss evaluated on a batch of samples cannot typically be expressed as a sum or average of losses evaluated at individual samples, which in turn requires new theoretical and methodological developments beyond standard treatments of supervised learning. In this paper, we advance this understanding of binary classification for complex performance measures by identifying two key properties: a so-called Karmic property, and a more technical threshold-quasi-concavity property, which we show is milder than existing structural assumptions imposed on performance measures. Under these properties, we show that the Bayes optimal classifier is a threshold function of the conditional probability of positive class. We then leverage this result to come up with a computationally practical plug-in classifier, via a novel threshold estimator, and further, provide a novel statistical analysis of classification error with respect to complex performance measures.
\end{abstract}

\section{Introduction}
\label{sec:intro}
Binary classification, with the goal of predicting a binary response given input features, is perhaps {\em the} classical problem in machine learning, with wide ranging applications. A key ingredient in binary classification is a \emph{performance measure}, that quantifies how well a given classifier fits the data. While the performance measure of \emph{accuracy} has been the predominant focus of both theory and practice, it has severe limitations in many practical settings, such as imbalanced classes, and where different types of errors made by the classifier have different costs~\cite{gu2009evaluation, wallace2011class}. Accordingly, practitioners in applied machine learning settings such as information retrieval and medical diagnosis have developed complex performance metrics that capture important trade-offs between different types of errors; we have collated a few instances in Table \ref{tab:metrics}. A key complication with many complex classification performance metrics, such as the F-measure \citep{manning2008introduction} and Harmonic Mean \citep{kennedy2009learning}, is that they cannot be decomposed into the sum or average of individual losses on each sample. Even the simple performance measure of precision --- the fraction of correct positive predictions, among the set of positive predictions --- is not a sum of individual losses on each sample. Thus, the standard theoretical and practical treatments of  supervised learning, such as standard empirical risk minimization that minimizes the empirical expectation of a loss evaluated on a single random example, are not applicable.

This practical reality has motivated research into effective and efficient algorithms tailored to complex non-decomposable performance measures. One class of approaches extend standard empirical risk minimization to this non-decomposable setting, which often relies on strong assumptions on either the form of the classifiers, such as requiring linear  classifiers~\cite{narasimhan2015optimizing}, or restricted to specific performance measures such as F-measure~\cite{parambath2015theory}. An alternative approach is the plug-in estimator, where we first derive the form of the Bayes optimal classifier, estimate the statistical quantities associated with the Bayes optimal classifier, and finally ``plug-in'' the sample estimates of the population quantities to then obtain the overall estimate of the Bayes optimal classifier. In particular, for many complex performance metrics, the Bayes optimal classifier is simply a thresholding of the conditional probability of the positive class~\citep{koyejo2014consistent, narasimhan2014statistical}, so that the plug-in estimator requires (a) an estimate of the conditional probability, and (b) the associated threshold. Plug-in methods have been of particular interest in non-parametric functional estimation as they typically require weaker assumptions on the function class and are often easy to implement.

In this paper, we seek to advance our understanding and practice of binary classification under complex non-decomposable performance measures. We show that for a very broad class of performance measures, encompassing a large set of performance measures used in practice, the Bayes optimal classifier is simply a thresholding of the conditional probability of the response. Towards this general result, we identify two key properties that a performance measure could satisfy. The first is what we call a ``Karmic'' property that loosely has the performance measure be more sensitive to an increase in true positives and true negatives, and a decrease in false positives and false negatives. The second is a more technical property we call threshold-quasi-concavity, which in turn ensures the performance measure is well-behaved around an optimal threshold. As we show these properties are satisfied by performance metrics used in practice, and in particular, these conditions are milder than existing results that restrict either the structural form of the performance measures, or impose strong shape constraints such as particular monotonicities. 

Our general result has two main consequences, which we investigate further: one algorithmic, and the other for the analysis of classification error for general performance measures. As the algorithmic consequence, we leverage the derived form of the Bayes optimal classifier, and some additional general assumptions on the performance measures, to provide a tractable algorithm to estimate the threshold, which coupled with an estimator of the conditional probability, provides a tractable ``plug-in estimator'' of the Bayes optimal classifier. Towards the statistical analysis consequence, we provide an analysis of the excess classification error, but with respect to general non-decomposable performance measures, of the general class of plugin-estimators for our class of Bayes optimal classifiers. Our analysis of classification error rates for such plug-in classifiers depend on three natural quantities: the rate of convergence for the conditional probability estimate, the rate of convergence for the threshold estimate, and a measurement of noise in the data. For the last part, we extend margin or low-noise assumptions for binary classification with the accuracy performance measure to complex performance measures. \emph{Low noise} assumptions, proposed by \citet{mammen1999smooth} in the context of the accuracy performance measure, bounds the noise level in the neighborhood of the Bayes optimal threshold i.e. $\frac{1}{2}$ for standard classification. Under such a low-noise assumption, \citet{audibert2007fast} derive fast convergence rates for plug-in classification rules based on the smoothness of the conditional probability function. Similar margin assumptions have also been introduced for density level set estimation by \citet{polonik1995measuring}. We provide a natural extension of such a low-noise assumption, under which we provide explicit rates of convergence of classification error with respect to complex performance measures. We provide corollaries for both parametric and non-parametric instances of our general class of plugin-classifiers.

\begin{table*}[t]
\centering
\caption{Examples of evaluation metrics. {\em Notation:} $\text{TPR}=\frac{\tp}{\tp+\fn}; \text{TNR}=\frac{\tn}{\tn+\fp}$.}
\label{tab:metrics}
\vskip 0.15in
\begin{sc}
\begin{tabular}{lllll}
\hline
\abovespace\belowspace
Metric & Definition & Reference & $\cG(\bc)$ \\
\hline
\abovespace
Accuracy 			& \small $\tp+\tn$				& 						& \small $(1,0,0,1)\bc$\\
Arithmetic Mean (AM)        &\small  $(\text{TPR}+\text{TNR})/2$  & \citet{menon2013statistical}   & \small $(\frac{\bc_1}{\bc_1+\bc_3}+\frac{\bc_4}{\bc_2+\bc_4})/2$\\
Youden's Index    	&\small  $\text{TPR}+\text{TNR}-1$ 	& \citet{youden1950index} 	& \small $ \frac{\bc_1}{\bc_1+\bc_3}+\frac{\bc_4}{\bc_2+\bc_4}-1$ \\
$F_\beta$ 		&\small  $\frac{(1+\beta^2)\tp}{(1+\beta^2)\tp+\beta^2 \fn+\fp}$ & \citet{van1974foundation}  & \small $\frac{(1+\beta^2,0,0,0)\bc}{(1+\beta^2,1,\beta^2)\bc}$ \\
Linear-Fractional  	&\small  $\frac{a_1\tp+a_2\fp+a_3\fn+a_4\tn}{b_1\tp+b_2\fp+b_3\fn+b_4\tn}$ & \citet{koyejo2014consistent} & \small $ \frac{\baa^T\bc}{\bb^T\bc}$ \\
G-Mean 			&\small  $\sqrt{\text{TPR}\cdot \text{TNR}}$ & \citet{daskalaki2006evaluation}	&\small  $\sqrt{\frac{\bc_1\bc_4}{(\bc_1+\bc_3)(\bc_2+\bc_4)}}$ \\
Q-Mean 			&\small  $1-\sqrt{\frac{ (1-\text{TPR})^2+(1- \text{TNR})^2 }{2}}$ & \citet{kubat1997addressing}	& \small $1-\sqrt{\frac{( \frac{\bc_3}{\bc_1+\bc_3})^2+( \frac{\bc_2}{\bc_2+\bc_4})^2}{2}}$ \\
\belowspace
H-Mean 			&\small  $\frac{2}{1/\text{TPR}+1/\text{TNR}}$ & \citet{kennedy2009learning}& \small  $2\left/ \left( \frac{\bc_1+\bc_3}{\bc_1}+\frac{\bc_2+\bc_4}{\bc_4}\right)\right.$ \\
\hline
\end{tabular}
\end{sc}
\vskip -0.1in
\end{table*}

The rest of the paper is organized as below.
In Section 2 we introduce the problem and relevant notations. The characterization and properties of Bayes optimal classifier are derived in Section \ref{sec:bayes}. We discuss the algorithm for estimating the plug-in estimator in Section~\ref{sec:estimation}, and present the statistical convergence guarantee in Section~\ref{sec:fast_rate}. Applications of the derived rate for two special cases, Gaussian generative model and $\beta$-H\"{o}lder class conditional probability are presented in Section \ref{sec:example} where explicit convergence rates are provided. We conclude the paper in Section \ref{sec:conclusion}. Detailed proofs are deferred to the supplementary materials.

\section{Problem Setup and Preliminaries} 
\label{sec:prelim}

Binary classification entails predicting a binary label $Y \in \{\pm 1\}$ associated with a feature vector $X \in \cX\subset  \bR^d$. Such a a function mapping $f: \mathcal{X} \mapsto \{\pm 1\}$ from the feature space $\mathcal{X}$ to the labels $\{\pm 1\}$ is called a binary classifier.  Let $\Theta=\left\{f: \cX \to \{\pm 1\}\right\}$ denote a set of binary classifiers.  We assume $(X, Y)$ has distribution $\bP \in \mathcal{P}$, and let $\eta(x) :=\bP(Y=1|X=x)$ denote the conditional probability of the label $Y$ given feature vector $x$. 

A key quantity is the confusion matrix, that consists of four population quantities: true positives (TP), true negatives (TN), false positives (FP), and false negatives (FN). 

\begin{definition}[Confusion Matrix]\label{def:confusion}
For any classifier $f:\bR^d \mapsto \{\pm 1\}$, its confusion matrix is defined as $\bC(f,\bP) := [\tp(f,\bP),\fp(f,\bP),\fn(f,\bP),\tn(f,\bP)]\in [0,1]^4$, where:
\begin{equation}
\begin{split}
\tp(f,\bP) &=\bP (Y=+1,f(X)=+1) , \\
\fp(f,\bP) &=\bP (Y=-1,f(X)=+1), \\
\fn(f,\bP) &=\bP (Y=+1,f(X)=-1), \\
\tn(f,\bP) &= \bP(Y=-1,f(X)=-1).
\end{split}
\label{def:fundmetric}
\end{equation}
\end{definition}

Another key ingredient is the utility or performance measure $\mathcal{U}:\Theta \times \mathcal{P} \to \mathbb{R}$, that measures the performance of a classifier. In this paper, we focus on complex binary classification performance measures that can be expressed as a function of the confusion matrix. Formally, $\cU(f, \bP) = \cG(\bc(f,\bP))$.  When it is clear from context, we will drop the dependency of the distribution $\bP$ in $\bC$ and $\cU$. The confusion-matrix functions $\cG$ corresponding to popular performance measures are listed in Table~\ref{tab:metrics}. Given the performance measure $\mathcal{U}$, we are interested in the corresponding Bayes optimal classifier:
\begin{equation}
f^*=\arg\max_{f \in \Theta}\; \mathcal{U}(f, \bP).
\label{eq:fstar}
\end{equation}
Given any candidate classifier $f$, we are then interested in the excess risk $\cU(f^*,\bP)-\cU(f,\bP)$, which is the utility gap between a given classifier $f$ and the corresponding Bayes optimal.

\begin{assumption}[Karmic Performance Measure]
The confusion-matrix function $\cG$ corresponding to the performance measure $\cU$ is Lipschitz~continuous, and satisfies the condition that $\nabla \cG(\bC)^T(1,-1,-1,1)^T\ge C_B$, for some constant $C_B>0$. 
\label{ass:utility}
\end{assumption}
\vspace{-1em}
We term performance measures that satisfy the condition $\nabla \cG(\bc)^T(1,-1,-1,1)^T\ge C_B$ as ``Karmic measures'', since it guarantees a lower bound on the sensitivity of the performance measure in the direction of increasing true positives and true negatives, and decreasing false positives and false negatives. 
While our Karmic assumption slightly weakens the existing monotonicity assumption used in literature, it is worth pointing out that the analysis in~\cite{narasimhan2014statistical} requires not only monotonicity but also additional assumptions (Assumption B in~\cite{narasimhan2014statistical}). Assumption B assumes the existence and uniqueness of an optimal threshold, which turns out to be non-trivial to check. Our analysis on the threshold-quasi-concavity closes this gap.

Assumption~\ref{ass:utility} is satisfied if $\cG$ is strictly monotonically increasing with respect to $\tp,\tn$ and decreasing with respect to $\fp,\fn$. Such an assumption is natural in that one would typically prefer a classifier with higher $\tp$ for fixed $\tn$ and vice versa~\citep{narasimhan2015consistent}. This condition is satisfied by most metrics in common use e.g. for the $F_1$ measure, $\nabla \cG(\bc)^T(1,-1,-1,1)^T=\frac{4(\fp+\fn)}{(2\tp+\fp+\fn)^2}$ is strictly positive as long as the data is not fully separable.

\subsection{Related Work}
\paragraph{Representation of the Bayes Optimal Classifier.}
The Bayes optimal classifier under the accuracy metric is classically known to be a thresholding of the conditional probability of the response, with the threshold of $1/2$ (see e.g. \citet{devroye2013probabilistic}). This property of Bayes optimal classifier having the thresholded form is called the probability thresholding principle for binary classification by \citet{lewis1995evaluating}. Prior work has also shown that the thresholding principle, with a metric dependent threshold, for more complex specific measures such as F-measure~\citep{jansche2007maximum, zhao2013beyond}, Arithmetic Mean (AM)~\citep{menon2013statistical}, linear-fractional performance metrics~\citep{koyejo2014consistent}, and monotonic concave metrics~\citep{narasimhan2015optimizing}. 

\paragraph{Plug-in Classifiers for Complex Metrics.}
For Bayes optimal classifiers that have thresholded form, a line of work has devised plug-in classifiers that then estimate the threshold, and the conditional probability of response. For the AM metric, \citet{menon2013statistical} show that the threshold is simply the proportion of the positive class. For linear fractional functions, \citet{koyejo2014consistent} provide an implicit characterization of the optimal threshold, but the solution of which in turn requires the knowledge of the optimal classifier, which is unknown in practice. As a practical estimator, \citet{koyejo2014consistent} propose an exhaustive search for the threshold over all data samples, and show that the resulting algorithm is consistent, but for which non-asymptotic convergence rates are not known. \citet{narasimhan2014statistical} also note the importance of estimating the optimal threshold, but do not provide practical algorithms. As we show in Section~\ref{sec:bayes}, the empirical risk as a function of the threshold is in general neither convex nor concave. Hence, care must be taken to construct an optimization algorithm that guarantees convergence to the true threshold. 

\paragraph{Estimators designed for specific Utility Functions.}
Perhaps the most studied non-decomposable performance metric is the F-measure~\citep{nan2012optimizing,joachims2005support, zhao2013beyond}, with wide use in information retrieval and related areas, and for which researchers have developed tailored estimators \citep{nan2012optimizing,joachims2005support} as well as risk bounds for these estimators~\citep{ zhao2013beyond}. For instance, \citet{busa2015online} propose a scalable online F-measure estimator for large-scale datasets, with a root finding algorithm for the threshold update which exploits special properties of the F-measure. Similarly, for the Arithmetic Mean (AM) measure, \citet{menon2013statistical} design a consistent optimization scheme, based on a balanced classification-calibrated surrogate to AM. Unfortunately, these techniques are not easily extended to general complex performance metrics. 

\paragraph{Algorithms for General Classification Measures.}
\citet{joachims2005support} poses the classification problem as a structured prediction problem, and for linear classifiers, propose a structural SVM solver, but for which neither consistency nor explicit convergence rates are known.~\citet{kar2014online} proposes an online gradient descent algorithm which requires function classes that satisfy a uniform convergence property, which is difficult to verify apriori. Along similar lines, \citet{narasimhan2015optimizing} propose a stochastic gradient method, that involves a linearization of classification metric. Their proposed approach depends strongly on the assumption of a linear (or kernelized) classifier, and it is not obvious that the procedure can be extended to more complex non-linear function classes.

\section{The Bayes Optimal Classifier Revisited}
\label{sec:bayes}
In this section, we characterize the Bayes-optimal classifier for the broad class of Karmic performance measures, that satisfy Assumption~\ref{ass:utility}. We then show that with one additional assumption, we call threshold-quasi-concavity, the optimal threshold can be guaranteed to be unique. This result will be crucial for the design and analysis of our computationally efficient threshold finding procedure in Section~\ref{sec:estimation}.

Denote $\mu$ as the measure corresponding to the marginal distribution of $X$.
The utility is Frech\'{e}t differentiable, whose Frech\'{e}t derivative of $\cU$ may be computed as:
\bas{
[\nabla \util(f)]_x =& \nabla \cG(\bc(f))^T \cdot [\nabla \bc(f)]_x \\
=& \frac{1}{2} \left( \nabla \cG(\bc)^T(1,-1,-1,1)^T \eta(x) \right. \\
&\quad \left.-\nabla \cG(\bc)^T(0,-1,0,1)^T \right) d\mu(x)
}
From the Karmic measure Assumption~\ref{ass:utility}, we know that $\nabla \cG(\bc)^T(1,-1,-1,1)^T> 0$.
We define the ``Bayes critical set'' of $\utilb(f, \bP)$ for any $f \in \mathcal{F}$ as the set of instances where the utility has zero derivative:
\bas{
A_3(f)=\left\{x: \eta(x)=\frac{\nabla \cG(\bc)^T(0,-1,0,1)^T}{\nabla \cG(\bc)^T(1,-1,-1,1)^T} \right\}. 
}
For notational simplicity, we have omitted the dependency of $g_i$ on $\bc(f)$. Similarly, we will use $A_3^* := A_3(f^*)$ to denote the Bayes critical set. 


In this paper we focus on distributions where the critical set of $\utilb(f,P)$ satisfies $\bP(A_3(f)) = 0$. For instance, this is true for any distribution that satisfies the following assumption.
\begin{assumption}[$\eta$-continuity]
Let $\nu$ denote the probability measure that is associated with random variable $Z=\eta(X)=P(Y=1|X)$, then $\nu$ is absolutely continuous with respect to $\mu$. Furthermore, the density of $\eta(X)$, denoted by $p_{\eta}(\cdot)$, has full support on $[0,1]$, and is bounded everywhere.
\label{ass:eta_injective}
\end{assumption}
Absolute-continuity guarantees the existence of the density of $Z$. Armed with the above assumption on the conditional probability of the response, we can then characterize the Bayes optimal classifier as follows.
\begin{theorem}[Bayes Optimal Classifier as a Thresholding Function]
Suppose that $\cU$ is a performance measure that satisfies Assumption~\ref{ass:utility}, and that $\eta(X)$ satisfies Assumption~\ref{ass:eta_injective}. Let $f^*$ be the Bayes classifier with respect to $\cU$ and $\bC^*$ be its confusion matrix. Then, for all $x\in (A_3^*)^c$,
\ba{
f^*(x)=\sign{\eta(x)-\frac{\nabla \cG(\bc^*)^T(0,-1,0,1)^T}{\nabla \cG(\bc^*)^T(1,-1,-1,1)^T}}.
\label{eq:optcl}
}
\label{th:thres}
\end{theorem}
 \vspace{-1em}
\paragraph{Threshold of Bayes optimal classifier} 
For some performance measures, the optimal threshold reduces to an absolute constant; for instance it has the value of $1/2$ for the accuracy measure $\util(f, \bP) = \tp+\tn$ (see e.g. ~\cite{devroye2013probabilistic}).
In the general case however, the optimal threshold $\delta^*$ is a solution of the fixed point equation:
{\small
\bas{
(\nabla \cG(\bc^*)^T(0,-1,0,1)^T)/(\nabla \cG(\bc^*)^T(1,-1,-1,1)^T) = \delta^*,
}}
which is fixed point equation due to the dependency of $\bc^*$ on the threshold $\delta^*$. Theorem~\ref{th:thres} guarantees the existence of a solution to the above fixed point equation, but not its uniqueness. As we will show in Section~\ref{sec:fast_rate}, uniqueness can be achieved with some  additional regularity assumptions.


We note that Theorem~\ref{th:thres} only imposes a weak Karmic assumption on the performance measure, which as as stated in Section~\ref{sec:prelim}, is more general than even a simple strictly monotonicity assumption. In particular, it generalizes prior work such as \cite{koyejo2014consistent, menon2013statistical}, that impose more stringent assumptions (linear or linear fractional form of the measures, or strong monotonicity conditions).

We next briefly discuss why the critical set is crucial. Consider for instance the example studied in \citet{narasimhan2014statistical}: with domain $\cX=\{x_1,x_2,x_3\}$, a corresponding probability mass function $(0.25,0.5,0.25)$, and the conditional probability $\eta=(0.49,0.5,0.51)$. \citet{narasimhan2014statistical} show that for this setting, and for the case of the H-mean measure, there exist at least two deterministic Bayes optima: $(-1,1,-1)$ and $(1,-1,1)\}$, which can be seen to not have a thresholded form i.e. it cannot be expressed as a (signed) thresholding of the conditional probability. Our analysis reveals why this is the case.

From the threshold expression in \eqref{eq:optcl} from Theorem~\ref{th:thres}, the optimal threshold can be computed explicitly as $\frac{\nabla \cG(\bc^*)^T(0,-1,0,1)^T}{\nabla \cG(\bc^*)^T(1,-1,-1,1)^T} = \frac{1}{2}$. Thus, the Bayes critical set $A_3^*=\{x:\, \eta(x) = \frac{1}{2} \} = \{x_2\}$ has measure $P(X \in A_3)=P(X = x_2) = \frac{1}{2} >0$. It is clear that the Bayes optimal classifier may not take a thresholded form on the Bayes critical set. 

\subsection{Uniqueness of the Bayes Optimal Threshold.}
We are interested in characterizing mild conditions on the performance measure under which the fixed point equation characterizing the Bayes optimal threshold has a unique solution, under which case $P(A_3^*)=0$ (guaranteed by the $\eta$-continuity Assumption~\ref{ass:eta_injective}). 

The performance measure restricted to classifiers that are threshold functions of the conditional probability, can be rewritten as a function of the conditional probability $\eta$ and the threshold $\delta$.  
\begin{definition}
We define $\cV_\eta(\delta,\bP) := \cU(f_{\eta,\delta}, \bP)$ as the performance measure of any threshold classifier $f_{\eta,\delta}(x)=\sign{\eta(x)-\delta}$. Its arguments are the threshold $\delta$, and distribution $\bP$, while the subscript $\eta$ notes its dependence on the conditional probability $\eta$.
\end{definition}
We next introduce the definition of quasi-concavity, and the assumption of $\cV$ being strictly quasi-concave.
\begin{definition}
A function $f:\cX \to \bR$ is said to be quasi-concave if $\forall x,y\in \cX$, such that $f(x)\le f(y)$, it follows that $\innerprod{\nabla f(x)}{y-x} \ge 0$. We further say that $f$ is strictly quasi-concave if it is quasi-concave and its gradient only vanishes at the global optimum, i.e., $f(y) < \max_{x\in \cX} f(x)\Rightarrow \|\nabla f(y)\|>0$.
\end{definition}
\vspace{-0.5em}
Quasi-concave functions have super level sets are convex sets, and moreover by definition are unimodal i.e. have a unique maximal point.
\begin{assumption}(Threshold-Quasi-Concavity)
The threshold-classifier performance measure $\cV_\eta(\delta,\bP)$ is strictly quasi-concave for $\delta\in [0,1]$.
\label{ass:V_qc}
\end{assumption}

Assumption~\ref{ass:V_qc} seems abstract, but it entails that the performance measure is well-behaved as a function of the threshold. Moreover, it can be easily shown to hold for performance measures in practical use. We provide a proposition that shows that the assumption is satisfied for two important classes of performance measures: linear-fractional functions and concave functions.

\begin{proposition}
If Assumptions \ref{ass:utility}, \ref{ass:eta_injective} hold, and either: (a) $\cG$ is twice continuously differentiable and concave, or (b) $\cG$ is a linear fractional function $\cG(\bc) = \frac{\baa^T\bc}{\bb^T\bc}$ with $|\bb^T\bc| > 0$. Then $\cV_{\eta}(\delta,\bP)$ is strictly quasi-concave. 
\label{prop:quasi-concave}
\end{proposition}

\begin{theorem}
Under Assumption~\ref{ass:V_qc}, the fixed-point equation:
\ba{
\delta =\frac{\nabla \cG(\bC(f_\delta),\bP)^T(0,-1,0,1)^T}{\nabla \cG(\bC(f_\delta),\bP)^T(1,-1,-1,1)^T},
\label{eq:fix_point_delta}
}
where $f_\delta(x) = \sign{\eta(x)-\delta}$, has a unique fixed point $\delta^*\in (0,1)$. Hence the threshold in Theorem~\ref{th:thres} is uniquely defined.
\label{th:uniqueness}
\end{theorem}

Theorems~\ref{th:thres} and \ref{th:uniqueness} have two key consequences: first, we can use the representation to design plugin-estimators of the Bayes optimal classifier; second, it facilitates the statistical analysis for rates of convergence. We will discuss each of these two consequences in the following sections.


\section{Algorithmic Consequence: Estimation of the Threshold}
\label{sec:estimation}

Theorem~\ref{th:thres} shows that for Karmic performance measures, the Bayes optimal classifiers has the thresholded form as in Eq.~\eqref{eq:optcl}, and moreover under the threshold-quasi-concavity Assumption~\ref{ass:V_qc}, this threshold is unique. An immediate algorithmic consequence of this is to focus on plug-in classifiers that separately estimate the conditional probability, and the threshold. We present this plugin-classifier template in Algorithm~\ref{alg:twostepgd}. The template needs: (a) an estimator for conditional probability density $\eta(x)$, and (b) an estimator for the threshold. For the convenience of analysis, we divide the set of samples into two independent subsets: the conditional probability estimator is estimated using one subset, and the threshold is estimated using the other. 
\begin{algorithm}[h!]
\caption{Two-step Plug-in Classifier for General Metrics}
\begin{algorithmic}[1]
 \STATE {\bfseries Input:}  Training sample $\{X_i,Y_i\}_{i=1}^n$, utility measure $\cU$, conditional probability estimator $\hat{\eta}$, stepsize $\alpha$.
\STATE Randomly split the training sample into two subsets $\{X_i^{(1)},Y_i^{(1)}\}_{i=1}^{n_1}$ and $\{X_i^{(2)},Y_i^{(2)}\}_{i=1}^{n_2}$;
\STATE Estimate $\heta$ on $\{X_i^{(1)},Y_i^{(1)}\}_{i=1}^{n_1}$;
\STATE Estimate $\hat{\delta}$ with $\{X_i^{(2)},Y_i^{(2)}\}$;
\STATE {\bfseries Output:}  $\hat{f}(x)=\sign{\heta-\hat{\delta}}$.
\end{algorithmic}
\label{alg:twostepgd}
\end{algorithm}
In the coming subsections we discuss how to estimate the conditional probability and the threshold respectively.

\subsection{Estimation of Conditional Probability Function}
The estimation of the conditional probability of the response plays a crucial role in the success of Algorithm~\ref{alg:twostepgd}, but we emphasize that it is not the focus of our paper. In particular, this is a well studied problem, and numerous methods have been proposed for both parametric and non-parametric model assumptions on the conditional probability function.

In this section we briefly discuss some common estimators, and defer additional details to Section~\ref{sec:example}.

\paragraph{Parametric methods.}
In a classical paper, \citet{ng2002discriminative} compares two models of classification: one can either estimate $P(Y)$ and $P(X|Y)$ first, then get the conditional probability by Bayes rule (generative model approach); or directly estimate $P(Y|X)$ (discriminative model approach). The two approaches can also be related. In particular, if $P_{\theta_Y}(X|Y)$ belongs to exponential family, we have
\bas{
P_{\theta_Y}(X|Y)=h(x)\exp\left(\langle\theta_Y,\phi(X)\rangle-A(\theta_Y)\right),
}
where $\phi(X)$ is the set of sufficient statistics, $\theta_Y$ is the vector of the true canonical parameters, and $A(\theta)$ is the log-partition function. Using Bayes rule, we then have:
\bas{
P(Y=1|X) = \frac{1}{1+\exp\left( -\langle \theta_1-\theta_0,\phi(X) \rangle+c^* \right)}
}
where $c^*=A(\theta_0)-A(\theta_1)$. The conditional distribution can be seen to follow a logistic regression model, with the generative model sufficient statistics as the features, and the difference of the generative model parameters serving as the parameters of the discriminative model. A natural class of estimators for either the generative or discriminative models is based on Maximum likelihood Estimation (MLE). In Section~\ref{sec:example}, we derive the rate of convergence for the special case where the generative distributions are Gaussians with same covariances for both classes.

\paragraph{Non-parametric methods.}
One can also estimate $\eta(x) = P(Y = 1|X)$ non-parametrically, where a common model assumption is some form of smoothness on $\eta(x)$. One popular class of smooth functions is the following.
\begin{definition}[$\beta$-H\"{o}lder class]
Let $\beta>0$, denote $\floor*{\beta}$ the maximal integer that is strictly less than $\beta$. For $x\in \cX$ and any $\floor*{\beta}$-times continuously differentiable real-valued function $\eta$ on $\cX$, we denote by $\eta_x$ its Taylor polynomial of degree $\floor*{\beta}$ at point $x$,
\bas{
\eta_x(x') = \sum_{s\le \floor*{\beta}} \frac{(x'-x)^s}{s!}D^s\eta(x).
}
$\beta$-H\"{o}lder class is defined as the functions that satisfy, for $\forall x,x'\in \cX$,
\bas{
|\eta_x(x)-\eta_x(x')|\le C_\beta \|x-x'\|^\beta.
}
In particular, when $0\le \beta<1$, we have
$|\eta(x)-\eta(x')|\le C_\beta \|x-x'\|^\beta$
where $\beta>0$.
\label{def:beta_holder}
\end{definition}
We can then estimate $\eta(x)$ from this family of smooth functions via locally polynomial estimators \cite{audibert2007fast}, or kernel (conditional) density estimators \cite{jiang2017uniform} with a properly chosen bandwidth. 

\subsection{Estimation of the Threshold}
When $\cV_\eta$ is quasi-concave, a key consequence is that its gradient with respect to the threshold suffices to provide ascent direction information. We leverage this consequence, and summarize a simple binary search algorithm based on the sign of $\cV'_\eta(\delta,\bP)$ in Algorithm~\ref{alg:binary_search}.

\begin{algorithm}[h!]
\caption{Binary search for the optimal threshold}
\begin{algorithmic}[1]
\STATE {\bfseries Input:}  Training sample $\{X_i,Y_i\}_{i=1}^n$, utility measure $\cU$, conditional probability estimator $\hat{\eta}$, tolerance $\epsilon_0$.
\STATE $\delta_\ell=0; \delta_r = 1$;
\WHILE {$|\delta_\ell-\delta_r|\ge \epsilon_0$}
\STATE Evaluate $s=\sign{\cV_{\heta}'(\delta,\bP_n)}$;
\IF {$s\ge 0$} 
\STATE $\delta_\ell = \frac{\delta_\ell+\delta_r}{2}$; 
\ELSE 
\STATE $\delta_r = \frac{\delta_\ell+\delta_r}{2}$;
\ENDIF
\ENDWHILE
\STATE {\bfseries Output:}  $ \frac{\delta_\ell+\delta_r}{2}$.
\end{algorithmic}
\label{alg:binary_search}
\end{algorithm}
In the next section, we then analyze the rates of convergence for the excess generalization error of the plug-in classifier learned from Algorithm~\ref{alg:twostepgd}, and with threshold estimated via Algorithm~\ref{alg:binary_search}.

\section{Statistical Analysis Consequence: Rates of Convergence}
\label{sec:fast_rate}
We next analyze the convergence rate of the excess utility. As we will show, the rates of convergence depend on three quantities: the noise level of the data distribution, the convergence rate of the conditional probability function, and the convergence rate of the threshold. We start by introducing some assumptions. 

We assume that the estimator of the conditional probability of response satisfies the following condition.
\begin{assumption}
Let $\cS_n$ denote a sample set of size $n$, and $\eta_{\cS_n}$ denote the conditional probability estimator learnt from $\cS_n$.
Then, for some absolute constants $c_1,c_2>0$, the conditional probability estimator satisfies the following condition:
\bas{
\sup_{\cS_n} P(|\eta_{\cS_n}(x)-\eta(x)|\ge \epsilon) \le c_1\exp(-c_2 \, a_n \, \epsilon^2) \qquad \text{a.e.}
}
\label{ass:eta_concentration}
\end{assumption}

The convergence rate also depends on the noise in the training labels, which is typically captured via the probability mass near the Bayes optimal threshold. Here we generalize the classical margin assumption (sometimes also called low noise assumption) of \citet{audibert2007fast}, developed for the accuracy metric, to the case where the optimal threshold is not a fixed constant $\frac{1}{2}$:
\begin{assumption}
For some function $C_0(\delta^*)>0$ that depends on the threshold $\delta^*$, there exists an $\alpha\ge 0$ such that
\bas{
P_X(0<|\eta(X)-\delta^*|\le t )\le C_0(\delta^*) \, t^\alpha.
}
\label{ass:ma}
\end{assumption}

The assumption characterizes the behavior of the regression function in the vicinity of the optimal threshold $\delta^*$. The case $\alpha=0$ bounds the probability by a constant potentially larger than one, and is trivially satisfied. The other extreme case $\alpha=\infty$ is most advantageous for classification, since in this case the regression function $\eta$ is bounded away from the threshold. 

In cases where the threshold is not an absolute constant (such as $1/2$), it has to be estimated from data. We make the following assumption on its convergence rate.
\begin{assumption}
Given a conditional probability estimate $\heta$ learned from an independent data source, the estimator $\hat{\delta}_n$ of the threshold, from a sample set of size $n$, satisfies the following condition, for some absolute constant $c_3>0$:
\bas{
P\left( \left| \cU(\sign{\heta-\delta^*}) -\cU(\sign{\heta-\hat{\delta}}) \right| \ge b_n^{-1}\right) \le  n^{-c_3}.
}
\label{ass:conv_of_thres_finding}
\end{assumption}
\vspace{-2em}
Note that Assumption~\ref{ass:conv_of_thres_finding} allows the rate $b_n$ to in turn depend on $\heta$, or more specifically, $\bE_X |\eta(X)-\heta(X)|$. Moreover, it does not necessarily require that $\hat{\delta}$ converge to $\delta^*$, only that their corresponding utility function values be close. 

Armed with these largely notational assumptions, we can now provide the rate for the overall data-splitting two-step plug-in classifier described in Algorithm~\ref{alg:twostepgd}:
\begin{theorem}
Suppose Assumption~\ref{ass:utility} and \ref{ass:eta_injective} hold,
and further that Assumptions~\ref{ass:eta_concentration} and \ref{ass:conv_of_thres_finding} hold for some $\heta$ and $\hat{\delta}$. Let $\cU^*=\cU(\sign{\eta-\delta^*},\bP)$ be the Bayes optimal utility. 
If we split the data as $n_1=n_2=\frac{n}{2}$, then with probability greater than $1-n^{-c_4}$:
\bas{
\cU^*-\cU\left(\sign{\heta-\hat{\delta}},\bP\right) \le c_5 \max\left\{ a_n^{-\frac{1+\alpha}{2}},b_n^{-1} \right\}.
}
where $c_4,c_5>0$ are absolute constants.
\label{Thm:MainThm}
\end{theorem}
\subsection{Key Lemmas}
We provide a detailed proof of the theorem in the Appendix, but provide brief vignettes via some key lemmas that also provide some additional insight into the statistical analysis. A key tool when analyzing traditional binary classification is to turn the excess risk into an expectation of the absolute deviation of conditional probability from the threshold $\frac{1}{2}$. We show in the following lemma that a similar result holds with general optimal threshold:
\begin{lemma}
Let $\bc_n$ and $\bc^*$ be the vectorized confusion matrices associated with $f_n=\sign{\eta_n-\delta^*}$ and $f^*=\sign{\eta-\delta^*}$ respectively, where $\delta^*$ is the threshold for the Bayes optimal classifier. Denote $C_G := \nabla \cG(\bC^*)^T(1,-1,-1,1)$, and $C_H := \max_f \|\nabla^2\cG(\bC(f))\|_{op}$, where $\|\cdot\|_{op}$ refers to the operator norm of a matrix. If for some constant $c_6$, $a_n\ge c_6\left(\frac{C_H}{C_G\min\{\delta^*,1-\delta^*\}}\right)^2$, then
\bas{
\cG(\bc^*) - \cG(\bc_n) &\ge \, \frac{1}{2}C_G\bE[ |\eta-\delta^*|1(f_n\neq f^*)],\\
\cG(\bc^*) - \cG(\bc_n) &\le \, \frac{3}{2}C_G\bE [|\eta-\delta^*|1(f_n\neq f^*)].
}
\label{th:first_order_approx_G}
\end{lemma}
\vspace{-1em}
This lemma thus helps us control the excess utility via the error of the conditional probability estimator $\heta-\eta$. Armed with this result, and additionally using Assumption~\ref{ass:eta_concentration} on the convergence rate of the conditional probability estimator, we can then show that the excess utility converges at the rate $O(a_n^{-\frac{1+\alpha}{2}})$:
\begin{lemma}
Suppose that Assumptions~\ref{ass:eta_concentration} and \ref{ass:ma} are satisfied, and that the Bayes optimal classifier is $f^* = \sign{\eta - \delta^*}$. Then there exists a constant $c_7>0$ which depend on $\cG$ and $\bc(f^*)$, such that 
$\cU(\sign{\eta-\delta^*}) - \cU(\sign{\eta_n -\delta^*}) \le c_7 a_n^{-\frac{1+\alpha}{2}}.$
\label{th:conv_rate_deltastar}
\end{lemma}
Lemma~\ref{th:conv_rate_deltastar} describes the classification error rate when the optimal threshold is known. Stitching this together with Assumption~\ref{ass:conv_of_thres_finding} on the convergence rate of the threshold estimator can then be shown to yield the statement of Theorem~\ref{Thm:MainThm}.

\subsection{Risk Bound for the Plugin Classifier from Algorithm~\ref{alg:binary_search}}

Prior work on threshold estimation for plug-in classifiers have ranged over brute-force search~\cite{koyejo2014consistent} with no rates of convergence, level-set based methods~\cite{parambath2015theory} for the specific class of linear fractional metrics, and Frank-Wolfe based methods~\cite{narasimhan2015consistent} for the specific class of concave performance metrics. However these estimators, in addition to focus on specific performance metrics, are only able to achieve a convergence rate of  $O(\max\{\bE\|\heta(X)-\eta(X)\|_1, 1/\sqrt{n} \})$. This entails that even if when the conditional probability estimator has a fast convergence rate, the final convergence rate for these estimators will still be bounded by $O(1/\sqrt{n})$. In this section we show that our simple threshold search procedure in Algorithm~\ref{alg:binary_search} achieves a fast $O(1/n)$ or $O(1/a_n)$ rate of convergence by leveraging our analysis from Section~\ref{sec:bayes}.

\begin{lemma}
Assume that Assumptions~\ref{ass:utility}, \ref{ass:eta_injective}, \ref{ass:ma} hold, and that the confusion-matrix function $\cG$ corresponding to the performance measure $\cU$ satisfies the same conditions as in Proposition~\ref{prop:quasi-concave}. Let $\hat{\delta}$ denote the output of Algorithm~\ref{alg:binary_search} with sample size $n$ and tolerance $\tau=\frac{\log n}{n}$, and $\heta$ denote a conditional probability estimator satisfying Assumption \ref{ass:eta_concentration} obtained on an independent sample set of size $n$. Denoting $\tilde{n} = \min\{n,a_n\}$, we then have that the rate $b_n$ in Assumption~\ref{ass:conv_of_thres_finding} satisfies: $b_n = \frac{\log \tilde{n}}{\tilde{n}}$.
\label{th:binary_guarantee}
\end{lemma}

An immediate corollary then gives the excess risk for the plug-in classifier.
\begin{corollary}
Suppose Assumption~\ref{ass:V_qc} holds. If $\tau=\frac{\log n}{n}, n_1=n_2=\frac{n}{2}$, then there exist constants $c_8,c_{9}>0$, such that with probability at least $1-\min\{n,a_n\}^{-c_8}$,
\bas{
\cU(f^*,\bP) - \cU(\hat{f},\bP) \le c_{9} \max\left\{ \frac{\log n}{n},\frac{\log a_n}{a_n},a_{n}^{-\frac{1+\alpha}{2}}  \right\}.
}
\label{thm:final_gen_bound}
\end{corollary}
\vspace{-1em}

\section{Explicit Rates for Specific Conditional Probability Models}
\label{sec:example}

In this section, we analyze two special cases where we can achieve explicit rate of convergence for the conditional probability estimation. 
For the first example, we consider the Gaussian generative model. We will show that the rate of convergence for the excess utility obtained in Theorem~\ref{thm:final_gen_bound} is $O(\frac{\log n}{n})$ in this case. The second example is for non-parametric kernel estimators when the conditional probability function satisfies certain smoothness assumption. 

\subsection{Gaussian Generative Model}
Consider two Gaussian distributions with the same variance, without loss of generality we assume the covariance matrix is identity $I_d$ for both classes. 
We define an asymmetric mixture of two Gaussians indexed by the centers and mixing weights.
\ba{
&\bP_{\mu, \kappa}: \quad P(Y=1)= \kappa, \, P(Y=0)=1- \kappa, \nonumber \\
& X|Y=1\sim \cN \left( \frac{\mu}{2},I_d \right), X|Y=0\sim \cN \left(-\frac{\mu}{2}, I_d \right).
\label{eq:model}
}
As stated in Section~\ref{sec:estimation}, we can compute the conditional probability and show that it can be fitted with a logistic regression model. Next we present results related to the key quantities in Theorem~\ref{thm:final_gen_bound}: $a_n$ and $\alpha$.
\begin{lemma}
Model defined in Eq.~\eqref{eq:model} with maximum likelihood estimator satisfies Assumption~\ref{ass:eta_concentration} with $a_n=n$. 
\label{lem:gaussian_an}
\end{lemma}

The following lemma specifies the margin assumption parameter for the above model. 
\begin{lemma}
The Gaussian generative model defined as in Eq.~\eqref{eq:model}, satisfies Assumption~\ref{ass:ma} with $\alpha=1$.
\label{lem:margin_gaussian}
\end{lemma}
Combining this result with Theorem \ref{thm:final_gen_bound} gives us the following corollary.
\begin{corollary}
Assume Assumptions \ref{ass:utility}-\ref{ass:ma} hold, $\bP$ is generated from Eq.~\eqref{eq:model}. Let $\hat{f}$ be the output of Algorithm~\ref{alg:twostepgd} with $\hat{\eta}$ estimated by MLE of logistic regression. We have with probability tending to 1, 
$\cU(f^*,\bP) - \cU(\hat{f},\bP) = O\left(   \frac{\log n}{n} \right).$
\label{cor:gaussian_rate}
\end{corollary}
\vspace{-0.5em}
For Gaussian generative models, fast rates of $O(\frac{1}{n})$ are only known for 0-1 loss \citep{li2015fast}. The logarithm factor can be further removed under 0-1 loss, or other cases when the threshold is known, as one can apply Lemma~\ref{th:conv_rate_deltastar} with $\alpha=1$ and get exactly the same rate as in \citet{li2015fast}. Corollary~\ref{cor:gaussian_rate} generalizes this result for a much broader class of utility functions, when the threshold is unknown and estimated from data.

\subsection{$\beta$-H\"{o}lder Densities}
When the conditional probability function belongs to the $\beta$-H\"{o}lder class as defined in Definition~\ref{def:beta_holder}, we have the following lemma on the convergence rates of $\eta_n$.
\begin{lemma}
For $\beta$-H\"{o}lder conditional probability functions, there exists estimators such that Assumption~\ref{ass:eta_concentration} holds with $a_n=n^{\frac{2\beta}{2\beta+d}}$.
\label{lem:beta_an}
\end{lemma}
Examples of such estimators include locally polynomial estimators \cite{audibert2007fast}, or kernel (conditional) density estimators \cite{jiang2017uniform}. 
Combined with Theorem~\ref{thm:final_gen_bound} we have the following corollary.
\begin{corollary}
Assume Assumptions~\ref{ass:utility}-\ref{ass:ma} hold and $\bP$ be a distribution where $P(Y=1|X)$ belongs to $\beta$-H\"older class. With locally polynomial estimators~\cite{audibert2007fast} or kernel (conditional) density estimators \cite{jiang2017uniform}, we have:
$\cU(f^*,\bP) - \cU(\hat{f}) = O\left(    n^{-\frac{(\min\{\alpha,1\}+1)\beta}{2\beta+d}} \right).$
\label{cor:beta_smooth}
\end{corollary}
The convergence rate obtained in Corollary~\ref{cor:beta_smooth} is faster than $O(\frac{1}{\sqrt{n}})$ if $\beta>\max\{ \frac{d}{2\alpha},\frac{d}{2} \}$. It is worth pointing out that the fast rate is obtained via a trade-off between the parameter $\alpha$ and $\beta$: to have a very smooth conditional probability function $\eta$, i.e., a large value of $\beta$, it cannot deviate from the critical level very abruptly, hence $\alpha$ has to be small.


\section{Conclusion}
\label{sec:conclusion}

We study Bayes optimal classification for general performance metrics. We derive the form of the Bayes optimal classifier, provide practical algorithms to estimate this Bayes optimal classifier, and provide novel analysis of classification error with respect to general performance metrics, and in particular show our estimators are not only consistent but have fast rates of convergence. We also provide corollaries of our general results for some special cases, such as when the inputs are drawn from a Gaussian mixture generative models, or when the conditional probability function lies in a H\"{o}lder space, explicitly proving fast rates under mild regularity conditions.


\bibliography{biblio}
\bibliographystyle{plainnat}
\end{document}